\documentclass{article}

\PassOptionsToPackage{numbers, compress}{natbib}

\usepackage[final]{nips_2018}

\usepackage[utf8]{inputenc} 
\usepackage[T1]{fontenc}    
\usepackage{hyperref}       
\usepackage{url}            
\usepackage{booktabs}       
\usepackage{amsfonts}       
\usepackage{nicefrac}       
\usepackage{microtype}      

\usepackage{verbatim}

\usepackage{threeparttable}

\usepackage{graphicx}

\usepackage{enumitem}

\usepackage{amssymb}
\usepackage{amsmath}
\usepackage{amsthm}

\title{Scene Graph Parsing by Attention Graph}

\author{
  Martin Andrews\\
  Red Dragon AI\\
  Singapore\\
  \texttt{martin@reddragon.ai} \\
  \And
  Yew Ken Chia\\
  Red Dragon AI\\
  Singapore\\
  \texttt{ken@reddragon.ai} \\
  \And
  Sam Witteveen\\
  Red Dragon AI\\
  Singapore\\
  \texttt{sam@reddragon.ai} \\
}

\begin{document}

\maketitle

\begin{abstract}
%

Scene graph representations, which
form a graph of visual object nodes together with their attributes and relations, 
have proved useful across a variety of vision and language applications. 
Recent work in the area has used Natural Language Processing
dependency tree methods to automatically build scene graphs.

In this work, we present an `Attention Graph' mechanism that can be trained end-to-end, 
and produces a scene graph structure that can be lifted directly from the top layer of a 
standard Transformer model.
%

The scene graphs generated by our model achieve an F-score similarity of 
52.21\% to ground-truth graphs on the evaluation set using the SPICE metric, 
surpassing the best previous approaches by 2.5\%.  
\end{abstract}

\section{Introduction}


In recent years, there have been rapid advances in the capabilities of 
computer systems that operate at the intersection of visual images and Natural Language Processing -
including semantic image retrieval \cite{johnson2015image,vendrov2015order},
image captioning \cite{mao2014deep,li2015cs231n,donahue2015long,liu2017attention}, 
visual question answering \cite{antol2015vqa,zhu2016visual7w,andreas2016neural},
and referring expressions \cite{hu2016natural,mao2016generation,liu2017recurrent}.
%

As argued in \cite{johnson2015image}, and more recently \cite{N18-1037}, 
encoding images as scene graphs (a type of directed graph that encodes information 
in terms of objects, attributes of objects, and relationships between objects)
is a structured and explainable way of expressing the knowledge 
from both textual and imaged-based sources, 
and is able to serve as an expressive form of common representation. 
The value of such scene graph representations has already
been proven in a range of visual tasks, 
including semantic image retrieval \cite{johnson2015image}, 
and caption quality evaluation \cite{anderson2016spice}. 

One approach to deriving scene graphs from captions / sentences is to use NLP methods for dependency parsing.
These methods extend the transition-based parser work of \cite{kiperwasser2016simple},
to embrace more complex graphs \cite{qi2017arc}, 
or more sophisticated transition schemes \cite{N18-1037}.


Recently, an alternative to the sequential state-based models underlying transition-based parsers
has gained popularity in general NLP settings, 
with the Transformer model of \cite{vaswani2017attention} leading
to high performance Language Models \cite{radford2018language}, 
and NLP models trained using other, innovative, criteria \cite{devlin2018bert}.





In this paper, we suppliment a pre-trained Transformer model
with additional layers that enable us to `read off' 
graph node connectivity and class information directly.
This allows us to benefit from recent advances 
in methods for training Language Models, 
while building a task-specific scene graph creation model.
The overall structure allows our graph elements to be created `holistically',
since the nodes are output in a parallel fashion, rather than through 
stepwise transition-based parsing.
%

Based on a comparison with other methods on the same 
Visual Genome dataset \cite{krishnavisualgenome} 
(which provides rich amounts of region description - region graph pairs),
we demonstrate the potential of this graph-building mechanism.
%
%
%

\begin{figure}
  \includegraphics[width=1.0\linewidth]{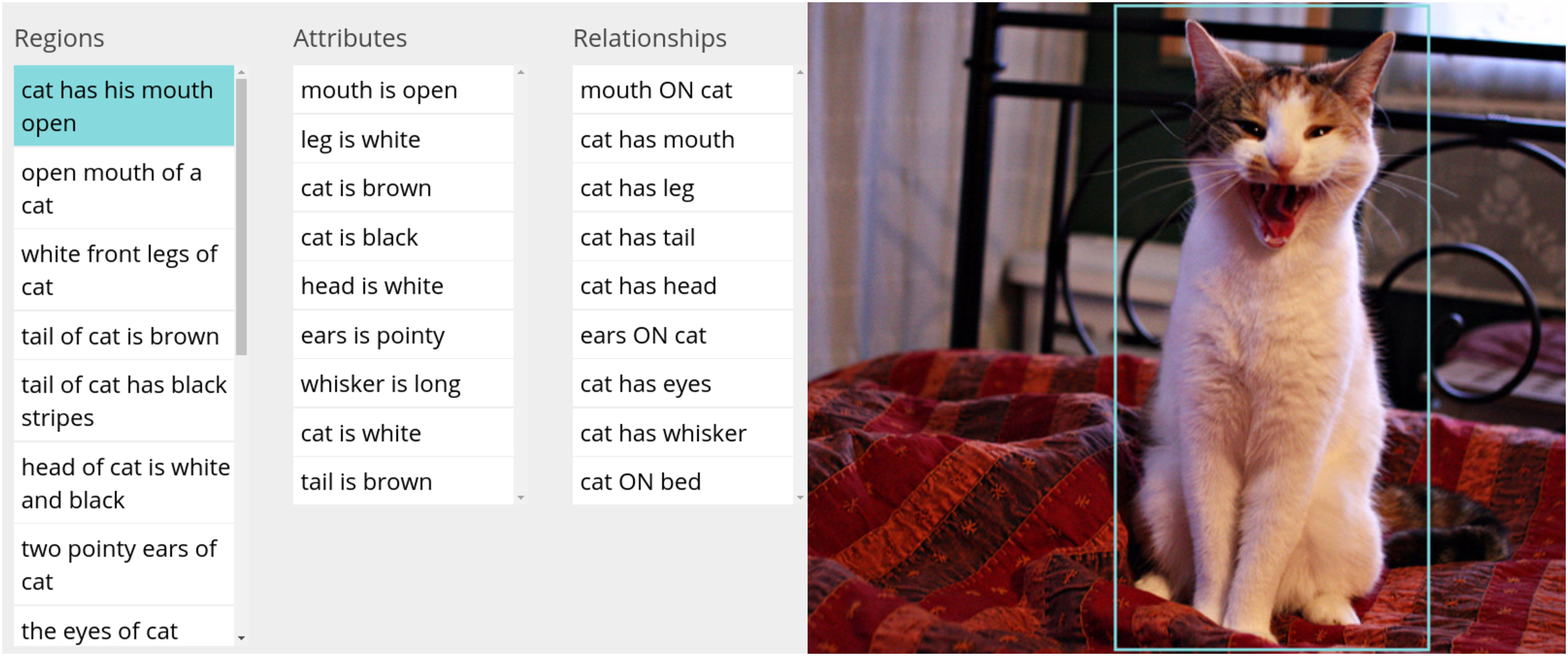}
  \caption{Example from data exploration site for \cite{krishnavisualgenome}. 
For this region, possible graph objects would be \texttt{\{cat, mouth\}}, 
attributes \texttt{\{brown$\leftarrow$cat, black$\leftarrow$cat, white$\leftarrow$cat\, open$\leftarrow$mouth\}}, and
relationships \texttt{\{cat$\leftarrow$has$\leftarrow$mouth, mouth$\leftarrow$ON$\leftarrow$cat\}}.
          }
  \label{fig:visual-genome}
\end{figure}

\section{Region descriptions and scene graphs}

Using the same notation as \cite{N18-1037}, 
there are three types of nodes in a scene graph: object, attribute, and relation.
Let $\mathcal{O}$ be the set of object classes, 
$\mathcal{A}$ be the set of attribute types, 
and $\mathcal{R}$ be the set of relation types.
Given a sentence $s$, our goal in this paper is to parse $s$ into a scene graph:
\begin{equation}
G(s) = \langle O(s), A(s), R(s) \rangle
\end{equation}
where $O(s) = \{o_1(s), \hdots, o_m(s)\}, o_i(s) \in \mathcal{O}$ is the set of object instances 
mentioned in the sentence $s$, 
$A(s) \subseteq O(s) \times \mathcal{A}$ is the set of attributes associated with object instances, 
and $R(s) \subseteq O(s) \times \mathcal{R} \times O(s)$ is the set of relations between object instances.

To construct the graph $G(s)$, we first create object nodes for every element in $O(s)$;
then for every $(o, a)$ pair in $A(s)$, we create an attribute node and add an unlabeled arc $o \leftarrow a$;
finally for every $(o_1, r, o_2)$ triplet in $R(s)$, we create a relation node and 
add two unlabeled arcs $o_1 \leftarrow r$ and $r \leftarrow o_2$.


\subsection{Dataset pre-processing and sentence-graph alignment}


We used the same dataset subsetting, training/test splits, 
preprocessing steps and graph alignment procedures as \cite{N18-1037}, 
thanks to their release of runnable program code\footnote{
Upon publication, we will also release our code, which includes some 
efficiency improvements for the preprocessing stage, as well as the models used.}.

\subsection{Node labels and arc directions}

In this work, we use six node types, which can be communicated using the CONLL file format:

\begin{enumerate}
\vspace{-1.0\topsep}
\item[\texttt{SUBJ}] The node label for an object in $\mathcal{O}$ 
(either standalone, or the subject of a relationship).  
The node's arc points to a (virtual) \verb+ROOT+ node

\item[\texttt{PRED}] The node label for a relationship $\mathcal{R}$,  
The node's arc points to \verb+SUBJ+

\item[\texttt{OBJT}] The node label for an object in $\mathcal{O}$ 
that is the grammatical object of a relationship, 
where the node's arc points to the relevant \verb+PRED+

\item[\texttt{ATTR}] The arc label from the head of an object node to the head of an attribute node.
The node's arc points to an object in $\mathcal{O}$  
of node type \verb+SUBJ+ or \verb+OBJT+

\item[\texttt{same}] This label is created for nodes whose label is a phrase. 
For example, the phrase ``in front of'' is a single relation node in the scene graph. 
The node's arc points to the node with which this node's text should be simply concatenated

\item[\texttt{none}] This word position is not associated with a node type, 
and so the corresponding model output is not used to create an arc

\end{enumerate}

\begin{figure}[ht]
  \centering
  \includegraphics[width=1.0\textwidth]{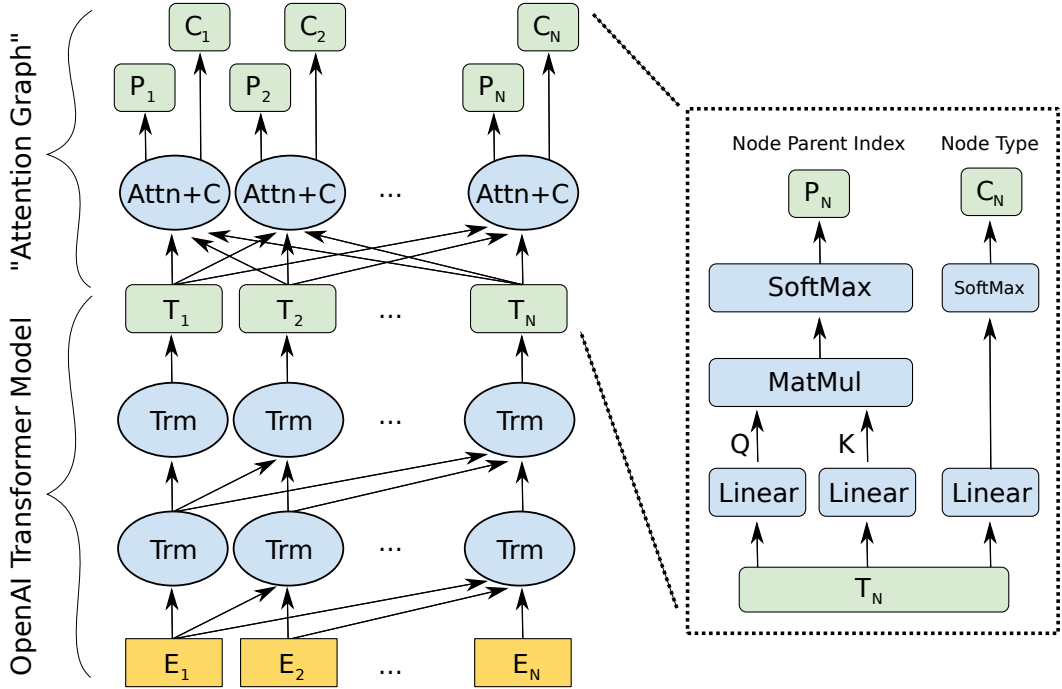}
    \caption{
      Model architecture illustrating Attention Graph mechanism
    }
    \label{fig:architecture}
\end{figure}

\section{Attention Graph Model}

The OpenAI Transformer \cite{radford2018language} Language Model was used as the foundation of 
our phrase parsing model (see Figure \ref{fig:architecture}).
This Transformer model consists of a Byte-Pair Encoded subword \cite{sennrich2015neural} embedding layer 
followed by 12-layers of ``decoder-only transformer with masked self-attention heads'' \cite{vaswani2017attention}, 
pretrained on the standard language modelling objective on a corpus of 7000 books.

The Language Model's final layer outputs were then fed in to a customised ``Attention Graph'' layer, 
which performed two functions : 
(i)~classifying the node type associated with each word; and 
(ii)~specifying the parent node arc required {\it from} that node.

The Attention Graph mechanism is trained using the sum of two cross-entropy loss terms
against the respective target node types and parent node indices, 
weighted by a factor chosen to approximately equalise the contributions to the total loss
of the classification and Attention Graph losses.
%
For words where the target node type is \verb+none+ (e.g: common conjunctions), 
the cross-entropy loss due to that node's connectivity is multiplied by zero, since its parent is irrelevant.

To convert a given region description to a graph, the 
BPE-encoded form is presented to the embedding layer $E_i$ in Figure \ref{fig:architecture}, 
and the node types and node arc destinations are read from $C_i$ and $P_i$ respectively.
No post-processing is performed : If the attention mechanism suggests an arc
that is not allowed (e.g.: an \verb+OBJT+ points to a word that is not a \verb+PRED+)
the arc is simply dropped.




\begin{table}[t]
  \parbox{.46\linewidth}{
    \centering
    \caption{SPICE metric scores for the Oracle 
      (using code released by \cite{N18-1037}) 
      and our method, 
      under the base assumptions, and also where the number of tuples 
      is bounded above by the number of potentially useful
      words in the region description
    }
    \label{tab:oracle}
    \begin{tabular}{lrrr}
      \toprule
      Parser & F-score                  & F-score  & F-score \\
             & reported                 & (our     & (limited \\
             & in \cite{N18-1037}       & tests)   & tuples) \\
      \midrule
      Attn. Graph     & & 0.5221 & {\bf 0.5750} \\
      (ours)  & &  &\\
      \midrule
      Oracle  & 0.6985 & 0.6630 & {\bf 0.7256} \\
      \bottomrule
    \end{tabular}
  }
  \hfill
  \parbox{.46\linewidth}{
    \centering
    \caption{SPICE metric scores between scene graphs parsed from region descriptions and ground truth region graphs 
       on the intersection of Visual Genome \cite{krishnavisualgenome} and MS COCO \cite{lin2014microsoft} validation set.}
    \label{tab:fscore}
    \begin{tabular}{lr}
      \toprule
      Parser & F-score \\
      \midrule
      Stanford \cite{schuster2015generating} & 0.3549 \\
      SPICE \cite{anderson2016spice} & 0.4469 \\
      Custom Dependency Parsing \cite{N18-1037} & 0.4967 \\
      \midrule
      Attention Graph (ours)  & {\bf 0.5221} \\
      \midrule
      Oracle (as reported in \cite{N18-1037}) & 0.6985 \\
      Oracle (as used herein) & 0.6630 \\
      \bottomrule
    \end{tabular}
  }
\end{table}

\section{Experiments}

We train and evaluate our scene graph parsing model on (a subset of) the Visual Genome \citep{krishnavisualgenome} dataset,
in which each image contains a number of regions, with each region being annotated with 
a region description and a (potentially empty) region scene graph.
Our training set is the intersection of Visual Genome and MS COCO \citep{lin2014microsoft} 
\verb+train2014+ set (34,027 images \& 1,070,145 regions), 
with evaluations split according to the MS COCO 
\verb+val2014+ set (17,471 images \& 547,795 regions).

We also tested the performance of the `Oracle' 
(an algorithmic alignment system between 
the region descriptions and the ground-truth graph tuples) 
- including a regime where the number of tuples was limited to the number of words, 
excluding \texttt{\{a, an, the, and\}}, 
in the region description.

The model $Q$ and $K$ vectors were of length $768$, consistent with the rest of the Transformer model.  
We use an initial learning rate of $6.25\times10^{-5}$ and Adam optimizer \citep{kingma2014adam} with $b_1$/$b_2$
of $0.9$/$0.999$ respectively.  Training was limited to 4 epochs (about 6 hours on a single Nvidia Titan~X).

\section{Results}

The $F_1$ scores given in Table \ref{tab:oracle} indicate that 
there might be significant room for improving the Oracle
(which, as the provider of training data, is an upper bound on any trained model's performance).  
However, examination of the remaining errors suggests that
an $F_1$ near 100\% will not be achievable because of 
issues with the underlying Visual Genome dataset.  
There are many instances where relationships are clearly stated 
in the region descriptions, where there is no corresponding graph fragment.  
Conversely, attributes don't appear to be region-specific, 
so there are many cases (as can be seen in Figure \ref{fig:visual-genome}) 
where a given object (e.g. `cat') has many attributes in the graph, 
but no corresponding text in the region description.

%
%
%

Referring to Table \ref{tab:fscore}, 
our Attention Graph model achieves a higher $F_1$ than previous work,
despite the lower performance of the Oracle used to train it\footnote{
This needs further investigation, since the Oracle results are 
a deterministic result of code made available by the authors of \cite{N18-1037}}.
The authors also believe that there is potential for further gains, 
since there has been no hyperparameter tuning, 
nor have variations of the model been tested 
(such as adding extra fully bidirectional attention layers).

\section{Discussion}

While the Visual Genome project is inspirational in its scope, 
we have found a number of data issues that put a limit on how
much the dataset can be relied upon for the current task.  
Hopefully, there are unreleased data elements that would allow
some of its perplexing features to be tidied up.

The recent surge in NLP benchmark performance has come through the use of 
large Language Models (trained on large, unsupervised datasets) 
to create contextualised embeddings for further use in downstream tasks.
As has been observed \cite{Ruder:NLP.ImageNet:2018}, 
the ability to perform transfer learning using NLP models 
heralds a new era for building sophisticated systems, 
even if labelled data is limited.  

The Attention Graph mechanism, as introduced here, also illustrates
how NLP thinking and visual domains can benefit from each other.
Although it was not necessary in the Visual Genome setting, 
the Attention Graph architecture can be further extended 
to enable graphs with arbitrary connectivity to be created. 
This might be done in several distinct ways, for instance
(a)~Multiple arcs could leave each node, using a multi-head transformer approach;
(b)~Instead of a SoftMax single-parent output $P_i$, 
multiple directed connections could be made using 
independent ReLU weight-factors; and
(c)~Potentially untie the correspondence that the Transformer 
has from each word to nodes, so that it becomes a Sequence-to-Graph model.
Using attention as a way of deriving structure is an interesting 
avenue for future work.

%

\newpage

\bibliographystyle{unsrtnat}
\bibliography{../nips2018_all}

\end{document}